\documentclass[journal]{IEEEtran}
\IEEEoverridecommandlockouts

\usepackage{cite}
\usepackage{amsmath,amssymb,amsfonts}
\usepackage{graphicx}
\usepackage{textcomp}
\usepackage{xcolor}

\usepackage{subcaption}
\usepackage{booktabs}
\usepackage[percent]{overpic}
\usepackage{colortbl}
\usepackage{array}
\usepackage{mdframed}
\usepackage{listings}

\definecolor{tableheader}{RGB}{45, 65, 95}
\definecolor{tablerowlight}{RGB}{245, 247, 250}
\definecolor{tablerowdark}{RGB}{230, 235, 242}
\definecolor{posgreen}{RGB}{34, 139, 34}
\definecolor{negred}{RGB}{178, 34, 34}
\definecolor{rulecomment}{RGB}{0, 100, 0}
\definecolor{rulenavy}{RGB}{20, 34, 66}
\newmdenv[
    backgroundcolor=tablerowlight,
    linecolor=tablerowdark,
    linewidth=0.6pt,
    roundcorner=3pt,
    innertopmargin=6pt,
    innerbottommargin=6pt,
    innerleftmargin=6pt,
    innerrightmargin=6pt
]{rulebox}
\lstdefinestyle{rulelisting}{
    basicstyle=\ttfamily\scriptsize,
    commentstyle=\color{rulecomment}\bfseries,
    emph={neural_rule_1,neural_rule_6},
    emphstyle=\color{rulenavy}\bfseries,
    backgroundcolor=\color{tablerowlight},
    frame=none,
    columns=fullflexible,
    keepspaces=true,
    showstringspaces=false,
    breaklines=true,
    breakatwhitespace=false,
    breakautoindent=false,
    breakindent=0pt,
    xleftmargin=0pt,
    aboveskip=0pt,
    belowskip=0pt
}

\usepackage{algorithm}
\usepackage{algorithmic}

\usepackage{setspace}
\usepackage{csquotes}

\usepackage[hidelinks]{hyperref}
\usepackage{orcidlink}

\DeclareUnicodeCharacter{03C4}{\ensuremath{\tau}}
\DeclareUnicodeCharacter{03B5}{\ensuremath{\epsilon}}
\DeclareUnicodeCharacter{03B3}{\ensuremath{\gamma}}
\DeclareUnicodeCharacter{03BB}{\ensuremath{\lambda}}

\newcommand{\ansrdt}{\textsc{Ansr-Dt}}
\newtheorem{proposition}{Proposition}
\def\BibTeX{{\rm B\kern-.05em{\sc i\kern-.025em b}\kern-.08em
    T\kern-.1667em\lower.7ex\hbox{E}\kern-.125emX}}

\begin{document}

\title{ANSR-DT: A Neuro-Symbolic Framework for Adaptive and Explainable Digital Twins}

\author{Safayat Bin Hakim, Muhammad Adil, Alvaro Velasquez, and Houbing Herbert Song%
\thanks{Safayat Bin Hakim and Houbing Herbert Song are with the Department of Information Systems, University of Maryland Baltimore County, Baltimore, MD 21250, USA (e-mail: shakim3@umbc.edu; songh@umbc.edu).}%
\thanks{Muhammad Adil is with the Department of Computer Science and Engineering, Texas Southern University, Houston, TX 77004, USA (e-mail: muhammad.adil@ieee.org).}%
\thanks{Alvaro Velasquez is with the Department of Computer Science, University of Colorado Boulder, Boulder, CO 80309, USA (e-mail: alvaro.velasquez@colorado.edu).}%
}
\maketitle

\begin{abstract}
Digital twins are increasingly used to monitor and optimize industrial systems, yet many existing frameworks remain difficult to interpret, slow to adapt, and limited in their ability to incorporate explicit domain knowledge. This paper presents \ansrdt, an adaptive neuro-symbolic framework that unifies temporal anomaly detection, symbolic reasoning, and reinforcement-learning-based decision support within a single digital twin pipeline. \ansrdt{} combines a CNN-LSTM model for multivariate pattern recognition with Prolog-based reasoning that converts learned signals into explicit rules, enabling transparent diagnoses and traceable decision paths. A PPO-based adaptation layer further refines operational responses under changing conditions while preserving interpretability. Experiments against eight baselines show that \ansrdt{} delivers competitive predictive performance together with stable rule extraction, scalable symbolic reasoning, and actionable explanations. Additional validation on the Skoltech Anomaly Benchmark (SKAB) further indicates that the framework transfers beyond synthetic settings. These findings position \ansrdt{} as a practical foundation for trustworthy, adaptive, and explainable industrial digital twins.

\hspace{1mm}

\noindent\textbf{Impact Statement---} Digital twins are increasingly deployed in industrial environments to monitor and optimize complex systems, yet many current implementations remain difficult to interpret and therefore hard to trust in human-in-the-loop settings. \ansrdt{} addresses this gap by combining neural prediction, symbolic reasoning, and adaptive control within a single framework that preserves traceable decision paths. By coupling anomaly detection with explicit rules and explanation-aware system states, the framework can support more transparent monitoring and more accountable operational responses in settings such as manufacturing and industrial process supervision. Its neuro-symbolic design is intended to improve collaboration between automated systems and human operators rather than replace human oversight. The open-source implementation further provides a practical basis for developing trustworthy and inspectable AI-enabled digital twins in safety-critical environments.

\end{abstract}


\vspace{-3mm}
\section{Introduction}
\label{sec:introduction}

Digital Twins (DTs), real-time digital replicas of physical systems, bridge the gap between physical and digital worlds to improve monitoring, decision-making, and optimization in complex scenarios such as smart factories, autonomous systems, and adaptable environments \cite{li2022digital, vohra2023digital, singh2022applications}. Despite their promise, traditional DT implementations struggle with adaptability and interpretability, fundamentally limiting trust in safety-critical industrial settings \cite{ogunsakin2023towards}. While Dynamic Digital Twins (DDTs) and Dynamic Data-Driven Application Systems (DDDAS) attempt to address these challenges through real-time data integration \cite{blasch2024dynamic}, they remain constrained by their inability to seamlessly incorporate human input and ensure transparent decision processes \cite{wang2024human}.

The core technical challenge lies in integrating real-time adaptability with interpretability---traditional models fail because they rely on static rule sets and lack continuous learning mechanisms \cite{aivaliotis2023methodology}. Current approaches suffer from \textbf{three critical limitations}: limited capacity for dynamic adaptation to real-time human input, opaque decision processes that lack transparency for human operators, and static symbolic representations that cannot evolve with changing operational conditions. Existing solutions either prioritize neural learning or symbolic reasoning, but rarely integrate them effectively to achieve both high performance and interpretability \cite{wan2024towards}.

To overcome these fundamental limitations, we propose \ansrdt, an Adaptive Neuro-Symbolic Learning and Reasoning Framework for digital twin technology. \ansrdt{} combines CNN-LSTM neural networks for dynamic event detection, Prolog-based symbolic reasoning for interpretable decision-making, and Proximal Policy Optimization (PPO) reinforcement learning \cite{schulman2017proximal} for continuous adaptation. This integration leverages attention mechanisms \cite{ham20203} to prioritize critical temporal patterns while maintaining logical clarity through symbolic reasoning \cite{kosasih2024review}, creating an adaptive system that addresses these critical limitations without sacrificing interpretability.

The principal contributions of this work are fourfold. First, we introduce a hybrid neuro-symbolic architecture that integrates deep learning and symbolic reasoning to improve decision-making in operational environments while addressing the interpretability gap in current digital twin implementations \cite{munir2023neuro}. Second, we develop a continuous learning mechanism through PPO-enhanced CNN-LSTM integration to support real-time adaptation to user preferences and environmental changes \cite{ma2024reinforcement}. Third, we provide a comprehensive empirical evaluation against eight baselines with statistical significance testing across three seeds, showing competitive classification performance together with quantitative scalability analysis up to 100 symbolic rules under real-time inference constraints. Finally, we release an open-source implementation to support reproducibility and to establish a practical foundation for future research on adaptive and interpretable digital twins.

The remainder of this paper proceeds as follows. Section~\ref{sec:background} examines related work and positions \ansrdt{} within the current research landscape. Section~\ref{sec:framework} details the three-layer architecture comprising physical sensor integration, neuro-symbolic processing, and reinforcement learning adaptation. Section~\ref{sec:implementation} describes the technical realization including synthetic data generation and component integration. Section~\ref{sec:results} presents comprehensive experimental evaluation demonstrating performance improvements and adaptability enhancements. Section~\ref{sec:discussion} analyzes the implications and limitations of our approach. Section~\ref{sec:conclusion} concludes with future research directions.

\vspace{-3mm}
\section{Background and Related Work}
\label{sec:background}
\vspace{-1mm}

Traditional digital twin implementations face persistent challenges that limit their effectiveness in dynamic industrial environments. Unlike static digital models \cite{walmsley2024adaptive}, contemporary DT systems require both adaptability to changing conditions and interpretability for human operators---requirements that existing approaches struggle to satisfy simultaneously. While recent work in neuro-symbolic AI \cite{schmidt2024systematic, zhou2022neuro, jacobneurosymbolic} and reinforcement learning \cite{ma2022consistency, ahmad2024accelerating} offers promising directions, these advances have not been systematically integrated into digital twin architectures tailored for industrial applications. This section reviews the four most relevant research fronts and identifies the convergence gap that \ansrdt{} addresses.

\vspace{-1mm}
\subsection{Neuro-Symbolic AI for Industrial Systems}

Several recent works pair neural and symbolic methods in industrial digital twin contexts, but each addresses only a subset of the capabilities that \ansrdt{} integrates. Schmidt et al.~\cite{schmidt2024systematic} systematically examined neuro-symbolic approaches for knowledge graph construction in manufacturing, identifying key integration patterns between subsymbolic and symbolic representations. Zhou et al.~\cite{zhou2022neuro} demonstrated practical neuro-symbolic AI deployment at Bosch, highlighting the importance of data foundations and systematic integration strategies. More recent efforts have explored complementary directions: Siyaev et al.~\cite{siyaev2023interaction} combined sequence-to-sequence models with symbolic program execution for natural-language interaction with aircraft maintenance digital twins, achieving 96.2\% query interpretation accuracy, but focused on NLP-driven interaction rather than real-time anomaly detection or adaptive control. Other recent work has proposed multi-layer knowledge graph models for manufacturing digital twins, improving quality metrics through semantic structuring, yet relied on KG-based reasoning without deep learning anomaly detection or RL-based adaptive response~\cite{ogunsakin2023towards}. Tian et al.~\cite{tian2024neural} introduced a temporal logic network encoding weighted signal temporal logic within neural layers for interpretable bearing fault diagnosis, but addressed single-component diagnosis only---lacking system-level DT orchestration and adaptive maintenance policies.

These works establish that effective neuro-symbolic integration requires careful attention to the bidirectional flow of information between neural and symbolic components---a principle central to \ansrdt's architecture. However, the surveyed literature does not provide an integrated digital twin framework that combines all three pillars considered here: neural detection, symbolic reasoning, and RL-based adaptation.

\vspace{-1mm}
\subsection{Explainable Anomaly Detection in Digital Twins}

Addressing the black-box problem in DT anomaly detection has attracted growing attention, as opaque decision processes undermine trust in safety-critical settings~\cite{renkhoff2024survey, bottjer2023review}. Kobayashi and Alam~\cite{kobayashi2024explainable} integrated SHAP, LIME, and PDP into a digital twin for remaining useful life prediction in nuclear energy systems, demonstrating how feature attribution enhances trust but remaining limited to post-hoc statistical attribution. Similarly, LSTM-Autoencoder-based detectors have been paired with SHAP for root-cause identification on industrial control system benchmarks, achieving strong precision but offering no mechanism for generating operational rules at runtime. More recent approaches have advanced toward integrated causal analysis within manufacturing DTs~\cite{wan2024towards}, yet still produce data-driven causal graphs rather than human-readable logical rules.

A recurring limitation across this area is reliance on post-hoc statistical explanation or data-driven causal analysis rather than runtime symbolic rule generation. \ansrdt's Prolog-based approach produces auditable, domain-grounded logical explanations during operation, and these rules directly condition the adaptive control policy.

\vspace{-1mm}
\subsection{Reinforcement Learning for Adaptive Digital Twins}

RL has been increasingly applied to digital twin adaptation, primarily for control and optimization rather than anomaly-aware decision-making. Khdoudi et al.~\cite{khdoudi2024deep} built a full-duplex DT for plastic injection molding combining supervised environment models with PPO and TD3 agents for autonomous process parameter optimization, but relied on opaque neural RL with no symbolic reasoning layer. Similarly, DQN and improved PPO variants have been applied to DT-driven reconfiguration planning and job-shop scheduling~\cite{ma2022consistency, ahmad2024accelerating}, focusing on optimization without symbolic domain knowledge or anomaly detection.

Across these works, RL in digital twins is mainly used for control, scheduling, or process optimization, with limited integration into symbolic reasoning or anomaly detection pipelines. \ansrdt{} bridges this gap by connecting CNN-LSTM-detected anomalies to PPO-driven responses mediated through interpretable Prolog rules.

\vspace{-1mm}
\subsection{Benchmark Validation with Interpretable Methods}

Evaluation on real industrial anomaly benchmarks (SWaT, WADI, SKAB, SMD, MSL/SMAP) is dominated by deep autoencoders and transformers, with only early steps toward symbolic or interpretable methods. Zhu et al.~\cite{zhu2025anomaly} mined variation-driven predicates and association rules with multiple minimum supports for ICS anomaly detection on SWaT and WADI, achieving high detection efficiency with low latency---the approach most comparable to \ansrdt's symbolic component, but without neural feature extractors or an adaptive remediation loop. Graph Attention Networks combined with GRU have been applied for inter-sensor dependency learning on SWaT and WADI, where ``interpretability'' was limited to attention-weight visualization rather than formal symbolic explanations. Correia et al.~\cite{correia2024online} provided a comprehensive survey of online multivariate time-series anomaly detection, critically analyzing standard benchmarks for fundamental flaws and explicitly identifying the combination of interpretability with neural detection as an unresolved challenge in the field.

To the best of our knowledge, evaluated systems on these benchmarks have not jointly fused formal symbolic reasoning, neural detection, and adaptive RL---the combination that \ansrdt{} provides through its CNN-LSTM detection layer, Prolog rule base updated from detector outputs, and PPO-driven adaptation.

\vspace{-1mm}
\subsection{Positioning of \ansrdt}

The reviewed 2023--2026 literature reveals a consistent pattern: existing work typically combines at most two of the three components (neural networks, symbolic models, RL) in digital twin settings. \ansrdt{} is positioned at the three-way intersection: it detects multivariate anomalies with CNN-LSTM, lifts detections into an executable Prolog rule base for human-readable reasoning, and uses PPO to learn anomaly-aware control policies guided by these symbolic rules---a combination not observed in the surveyed literature.

\section{Proposed Framework}
\label{sec:framework}

\subsection{\ansrdt{} Architecture Overview}

\begin{figure*}[!t]
    \centering
    \includegraphics[width=1.0\textwidth]{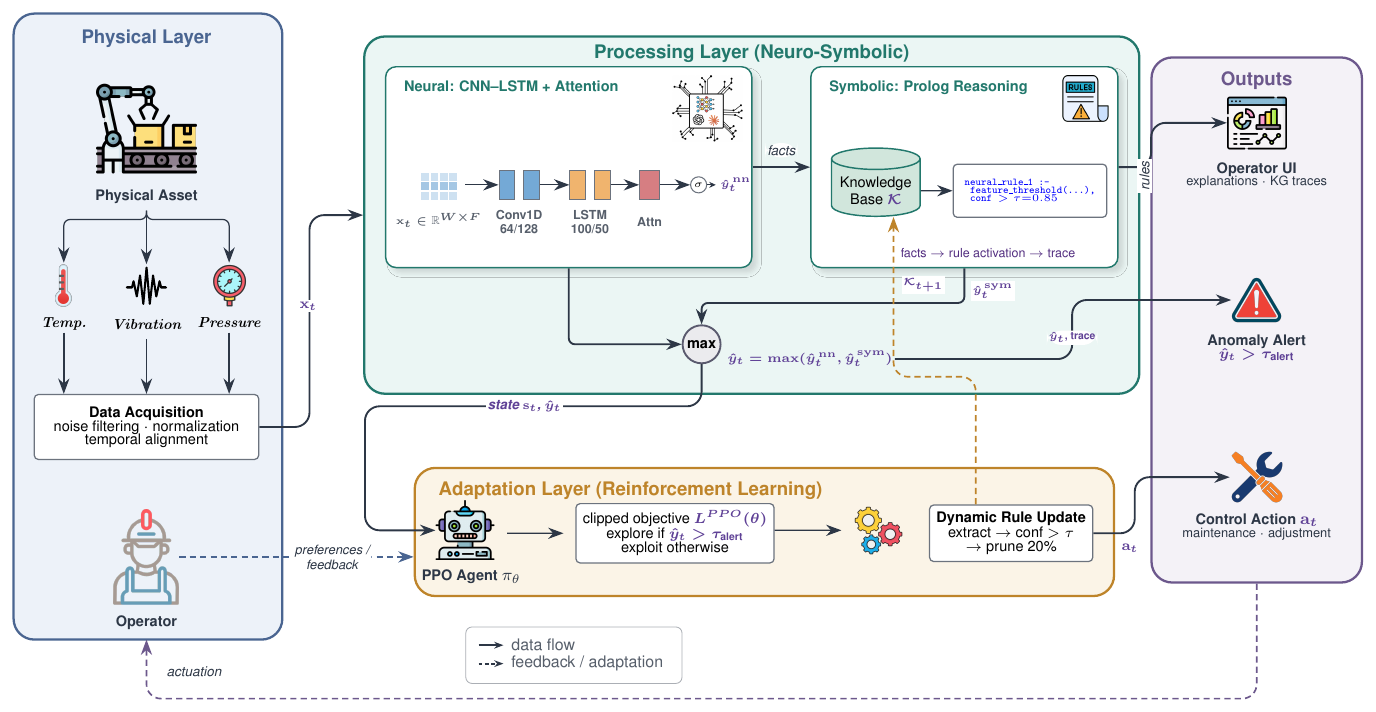}
    \caption{\small \ansrdt{} three-layer architecture: Physical Layer (sensor integration), Processing Layer (CNN-LSTM + symbolic reasoning), and Adaptation Layer (PPO + dynamic rule updating). Solid arrows: data flow; dashed arrows: feedback loops.}
    \label{fig:framework-complete}
    \vspace{-3mm}
\end{figure*}

\ansrdt{} integrates neural networks, symbolic reasoning, and reinforcement learning through a \textbf{three-layer architecture} (Fig.~\ref{fig:framework-complete}): a Physical Layer for sensor integration, a Processing Layer for neuro-symbolic reasoning, and an Adaptation Layer for continuous learning via PPO and dynamic rule updating.

\begin{figure*}[!t]
    \centering
    \includegraphics[width=.62\textwidth]{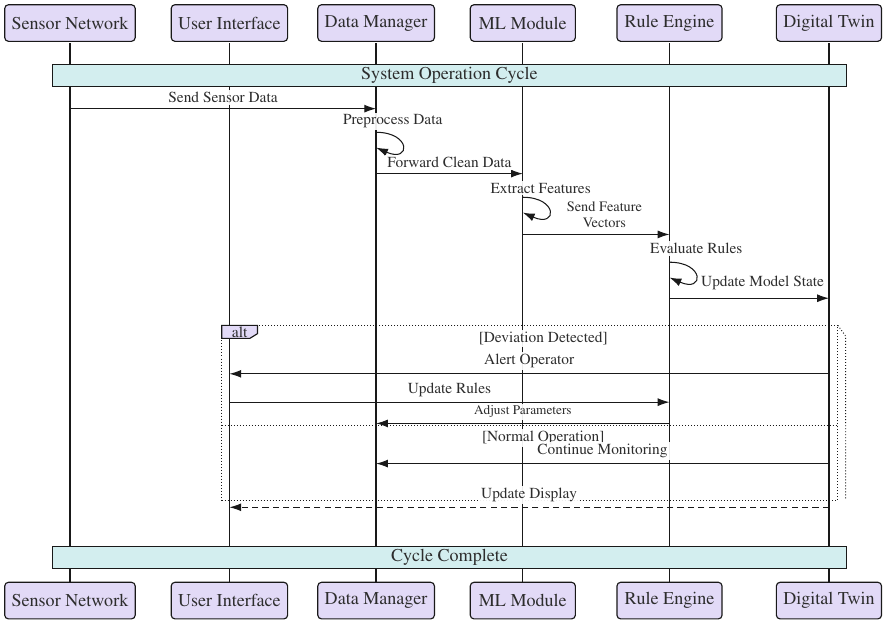}
    \caption{\small System operation sequence of the \ansrdt{} framework. The diagram illustrates the interaction between components such as the Sensor Network, Data Manager, ML Module, Rule Engine, Digital Twin, and User Interface for real-time time-series pattern extraction and adaptive operations.}
    \label{fig:sequence_diagram}
    \vspace{-2mm}
\end{figure*}

The Physical Layer preprocesses sensor data into standardized inputs. The Processing Layer extracts temporal patterns via CNN-LSTM and converts them into symbolic rules through Prolog-based inference. The Adaptation Layer refines policies via PPO and updates the symbolic knowledge base, forming a closed-loop system. The operational sequence is detailed in Fig.~\ref{fig:sequence_diagram}.

\subsection{Physical Layer: Sensor Integration}

The Physical Layer establishes the foundation for \ansrdt{} by connecting the digital twin to its physical counterpart through multiple sensor modalities:

\begin{itemize}
    \item \textbf{Temperature Sensors} monitor thermal variations in industrial equipment, detecting potential overheating or thermal deviations indicating degraded performance.
    \item \textbf{Vibration Sensors} measure mechanical oscillations, providing early warnings of component wear, imbalance, or structural degradation through characteristic frequency signatures.
    \item \textbf{Pressure Sensors} track fluid dynamics within systems, identifying potential leaks, flow restrictions, or anomalous pressure buildup.
\end{itemize}

Raw sensor data undergoes noise filtering, normalization, and temporal alignment to synchronize multi-sensor streams \cite{chen2021practical} before entering the Processing Layer.

\vspace{-2mm}
\subsection{Processing Layer: Neural-Symbolic Reasoning}
\vspace{-1mm}
The Processing Layer combines deep learning for temporal pattern extraction with symbolic reasoning for interpretable decision-making.

\subsubsection{Neural Component: CNN-LSTM Architecture}

The neural component employs a CNN-LSTM architecture: two Conv1D layers (64 and 128 filters, kernel size 3) with batch normalization and max-pooling extract spatial features, followed by two LSTM layers (100 and 50 units) for temporal dependencies. An attention mechanism prioritizes critical time steps. The model totals 189,481 parameters. The output is a per-sample anomaly probability $\hat{y} = \sigma(\mathbf{W}_o \mathbf{h}_{\text{att}} + b_o)$, where $\mathbf{h}_{\text{att}}$ is the attention-weighted LSTM output.

\subsubsection{Symbolic Component: Rule-Based Reasoning}

The symbolic engine translates neural outputs into Prolog-compatible rules \cite{sharma2024unsupervised}, maintaining a knowledge base of system states and operational constraints in first-order predicate logic. For each test sample, the neural model's prediction $\hat{y}_i$ and intermediate activations are analyzed to extract candidate rules $r$ with associated confidence scores. Rules are admitted to the knowledge base when:
\begin{equation}
    R_{\text{new}} = \{r \mid \text{conf}(r) > \tau, r \in \mathcal{F}(M_{\theta}, \mathcal{D})\}
    \label{eq:rule_extraction}
\end{equation}
where $\mathcal{F}$ extracts rules from model $M_\theta$ on data $\mathcal{D}$, and $\tau = 0.85$ is the confidence threshold. Rules encode two types of conditions: \textit{threshold-based} (e.g., {\small\texttt{feature\_threshold(vibration, \_, low)}}) comparing sensor values against learned thresholds, and \textit{gradient-based} (e.g., {\small\texttt{feature\_gradient(temperature, \_, high)}}) capturing rates of change. Body-level deduplication ensures that semantically identical rules extracted across different epochs are merged, retaining the highest-confidence version \cite{problog2}.

Operationally, the symbolic layer follows a compact rule lifecycle: neural predictions and selected feature trends are first converted into symbolic facts, candidate rules are extracted from recurrent high-confidence patterns, redundant rule bodies are merged, and only confidence-qualified rules are retained in the knowledge base. When multiple candidates overlap, confidence and specificity determine precedence, while periodic pruning keeps the rule base compact enough for real-time use without discarding the most informative conditions.

\subsection{Adaptation Layer: Reinforcement Learning}

The Adaptation Layer uses Proximal Policy Optimization (PPO) \cite{schulman2017proximal} to refine control policies through a clipped objective:

{\small
\begin{multline}
L^{PPO}(\theta) = \mathbb{E}_{t} \Big[
\min \Big( r_{t}(\theta) A_{t}, \\
\text{clip}(r_{t}(\theta), 1-\epsilon, 1+\epsilon) A_{t} \Big) - \beta H[\pi_\theta] \Big]
\label{eq:ppo_updated}
\end{multline}}

\noindent where $r_t(\theta) = \pi_\theta(a_t|s_t) / \pi_{\theta_{\text{old}}}(a_t|s_t)$ is the policy ratio, $A_t$ is the generalized advantage estimate, $\epsilon = 0.2$ constrains update magnitude, and $\beta = 0.01$ controls entropy regularization. The system employs context-aware exploration:
\begin{equation}
\pi(a|s) = \begin{cases}
\pi_{\text{explore}}(a|s) & \text{if deviation detected} \\
\pi_{\text{exploit}}(a|s) & \text{otherwise}
\end{cases}
\label{eq:exploration}
\end{equation}

\noindent switching to stochastic exploration when symbolic reasoning detects anomalies, and defaulting to deterministic exploitation during normal operation \cite{ahmad2024accelerating}. The Adaptation Layer also continuously updates the symbolic knowledge base with rules derived from RL outcomes, creating a feedback loop between policy learning and symbolic reasoning. Importantly, PPO refines the downstream control policy rather than the anomaly classifier itself: the classification results in Section~\ref{sec:results} arise from the neural-symbolic inference path, whereas PPO is evaluated through adaptation-oriented measures such as reward shaping, explained variance, and policy refinement under changing operating conditions.

\subsection{Neuro-Symbolic Integration Loop}

Algorithm~\ref{alg:integration} formalizes the end-to-end inference and adaptation loop that integrates all three layers. For each incoming sensor window, the neural component produces an anomaly probability; the symbolic engine evaluates rules against the current state; the fused prediction combines both signals; and the PPO agent selects a control action. Periodically, new rules are extracted from neural activations and the knowledge base is pruned.

In practical terms, a single inference cycle begins with a multivariate sensor window entering the CNN-LSTM, which produces an anomaly score and intermediate temporal features. These outputs are translated into symbolic facts describing threshold states and short-term trends, after which the Prolog engine activates matching rules and returns a symbolic anomaly signal together with the reasoning trace. The fused decision is then used both for anomaly reporting and, when necessary, for selecting a corrective action through PPO, allowing the system to pair predictive detection with a traceable operational response.

\begin{algorithm}[t]
\caption{ANSR-DT Neuro-Symbolic Integration}
\label{alg:integration}
\small
\begin{algorithmic}[1]
\REQUIRE Sensor window $\mathbf{x}_t \in \mathbb{R}^{W \times F}$, knowledge base $\mathcal{K}$, policy $\pi_\theta$
\ENSURE Fused prediction $\hat{y}_t$, control action $\mathbf{a}_t$, updated $\mathcal{K}$
\STATE $\hat{y}_t^{\text{nn}} \gets f_{\text{CNN-LSTM}}(\mathbf{x}_t)$ \COMMENT{Neural prediction}
\STATE $\mathbf{s}_t \gets \textsc{ToFacts}(\mathbf{x}_t, \hat{y}_t^{\text{nn}})$ \COMMENT{Symbolic facts}
\STATE $\hat{y}_t^{\text{sym}} \gets \textsc{PrologQuery}(\mathcal{K}, \mathbf{s}_t)$ \COMMENT{Rule eval.}
\STATE $\hat{y}_t \gets \max(\hat{y}_t^{\text{nn}},\; \hat{y}_t^{\text{sym}})$ \COMMENT{Fused score}
\IF{$\hat{y}_t > \tau_{\text{alert}}$}
    \STATE $\mathbf{a}_t \gets \pi_\theta(\cdot | \mathbf{s}_t)$ stochastic \COMMENT{Explore}
\ELSE
    \STATE $\mathbf{a}_t \gets \pi_\theta(\cdot | \mathbf{s}_t)$ deterministic \COMMENT{Exploit}
\ENDIF
\IF{extraction interval reached}
    \STATE $R_{\text{cand}} \gets \textsc{ExtractRules}(f_{\text{CNN-LSTM}}, \mathbf{x}_t)$
    \STATE $\mathcal{K} \gets \mathcal{K} \cup \{r \in R_{\text{cand}} : \text{conf}(r) > \tau\}$
    \STATE $\mathcal{K} \gets \textsc{Prune}(\mathcal{K}, \text{strategy})$ \COMMENT{Prune 20\%}
\ENDIF
\RETURN $\hat{y}_t, \mathbf{a}_t, \mathcal{K}$
\end{algorithmic}
\end{algorithm}

The fusion at line~4 uses a max-combination rather than a weighted average or learned combiner. This preserves a transparent safety-oriented decision rule: a sample is escalated whenever either the neural detector or the symbolic engine produces a high anomaly score. In practice, this choice tends to inherit the recall of the more sensitive component and may trade some precision for coverage, but it also ensures that a high-confidence symbolic rule can surface an event even when the neural score remains only moderate. Rule pruning (line~13) supports three strategies: confidence-based (remove lowest-confidence rules), LRU (least recently used), and LRA (least recently activated), each removing 20\% of rules per cycle while maintaining activation coverage (Section~\ref{sec:scalability}).

\subsection{Theoretical Properties}
\label{sec:theory}

Three properties characterize the fused inference of Algorithm~\ref{alg:integration} and connect directly to the empirical behavior in Section~\ref{sec:results}.

\begin{proposition}[Fusion recall guarantee]
\label{prop:fusion}
Let $D_{\mathrm{nn}}$ and $D_{\mathrm{sym}}$ be the sets of samples whose neural and symbolic scores exceed $\tau_{\text{alert}}$. The max-fused detector flags exactly $D_{\mathrm{nn}} \cup D_{\mathrm{sym}}$; consequently $R_{\text{fused}} \geq \max(R_{\mathrm{nn}}, R_{\mathrm{sym}})$ and $\mathrm{FP}_{\text{fused}} \leq \mathrm{FP}_{\mathrm{nn}} + \mathrm{FP}_{\mathrm{sym}}$.
\end{proposition}
\noindent\emph{Proof sketch.} $\max(\hat{y}^{\mathrm{nn}}, \hat{y}^{\mathrm{sym}}) > \tau_{\text{alert}}$ iff either score exceeds it, so detections form the union; recall is monotone under set inclusion on the positives, and fused false positives are contained in the union of the components' false positives. \hfill$\square$

Proposition~\ref{prop:fusion} formalizes the safety-oriented design: recall can never fall below the better component, while the precision cost is bounded by the weaker component's false alarms---a trade-off that becomes visible precisely when one detector generalizes poorly (Section~\ref{sec:skab}).

\begin{proposition}[Inference complexity]
\label{prop:complexity}
One cycle of Algorithm~\ref{alg:integration} costs $O(C_{\mathrm{nn}} + |\mathcal{K}| \, c_{\max})$, where $C_{\mathrm{nn}}$ is the fixed CNN-LSTM forward cost and $c_{\max}$ the maximum conditions per rule, i.e., symbolic overhead is linear in the rule count.
\end{proposition}
\noindent\emph{Proof sketch.} Each rule body is a conjunction of at most $c_{\max}$ ground facts evaluated once per window; no recursion is used. \hfill$\square$ This matches the measured latency growth from $1.80$\,ms at 5 rules to $5.33$\,ms at 100 rules (Section~\ref{sec:scalability}).

\begin{proposition}[Bounded knowledge base]
\label{prop:bounded}
If each extraction cycle admits at most $m$ rules and pruning removes a fraction $\rho$ of $\mathcal{K}$ per cycle (Eq.~\ref{eq:rule_update}), then $|\mathcal{K}_t| \leq \max\!\big(|\mathcal{K}_0|,\; \tfrac{(1-\rho)\,m}{\rho}\big)$ for all $t$, with geometric convergence to the fixed point.
\end{proposition}
\noindent\emph{Proof sketch.} $|\mathcal{K}_{t+1}| \leq (1-\rho)(|\mathcal{K}_t| + m)$ is an affine contraction with rate $1-\rho < 1$. \hfill$\square$ For $\rho = 0.2$ the rule base is thus bounded by $4m$ regardless of run length, ensuring bounded memory and, via Proposition~\ref{prop:complexity}, bounded latency.

\section{Implementation}
\label{sec:implementation}

This section details the technical realization of \ansrdt, emphasizing methodologies for synthetic data generation, component development, and system integration. The complete implementation, including documentation and example scripts, is available at {\renewcommand{\UrlFont}{\bfseries\fontfamily{pcr}\selectfont\small}\url{https://github.com/sbhakim/ansr-dt}}.

\subsection{Synthetic Data Generation and Rationale}

To evaluate \ansrdt{} in controlled yet realistic conditions, we developed a \textbf{synthetic data generation module} that simulates industrial manufacturing scenarios. This approach offers perfect ground truth labeling, systematic testing capabilities, and reproducibility for comparative evaluation. The synthetic environment allows validation of system performance without the privacy and security concerns associated with sensitive industrial data, while enabling precise control over the types and frequencies of operational deviations injected for testing.

The dataset comprises 5,000 samples at 5-minute intervals with seven variables: three sensor readings (temperature, vibration, pressure), operational hours, efficiency index, system state (categorical), and performance score. Key events are injected at 5\% rate across four types: operational threshold violations, maintenance events, state transitions, and interaction deviations. Inter-sensor correlations (0.3--0.5) and Savitzky-Golay smoothing \cite{chen2021practical} ensure realistic temporal characteristics.

\begin{figure}[htbp]
    \centering
    \includegraphics[width=\linewidth]{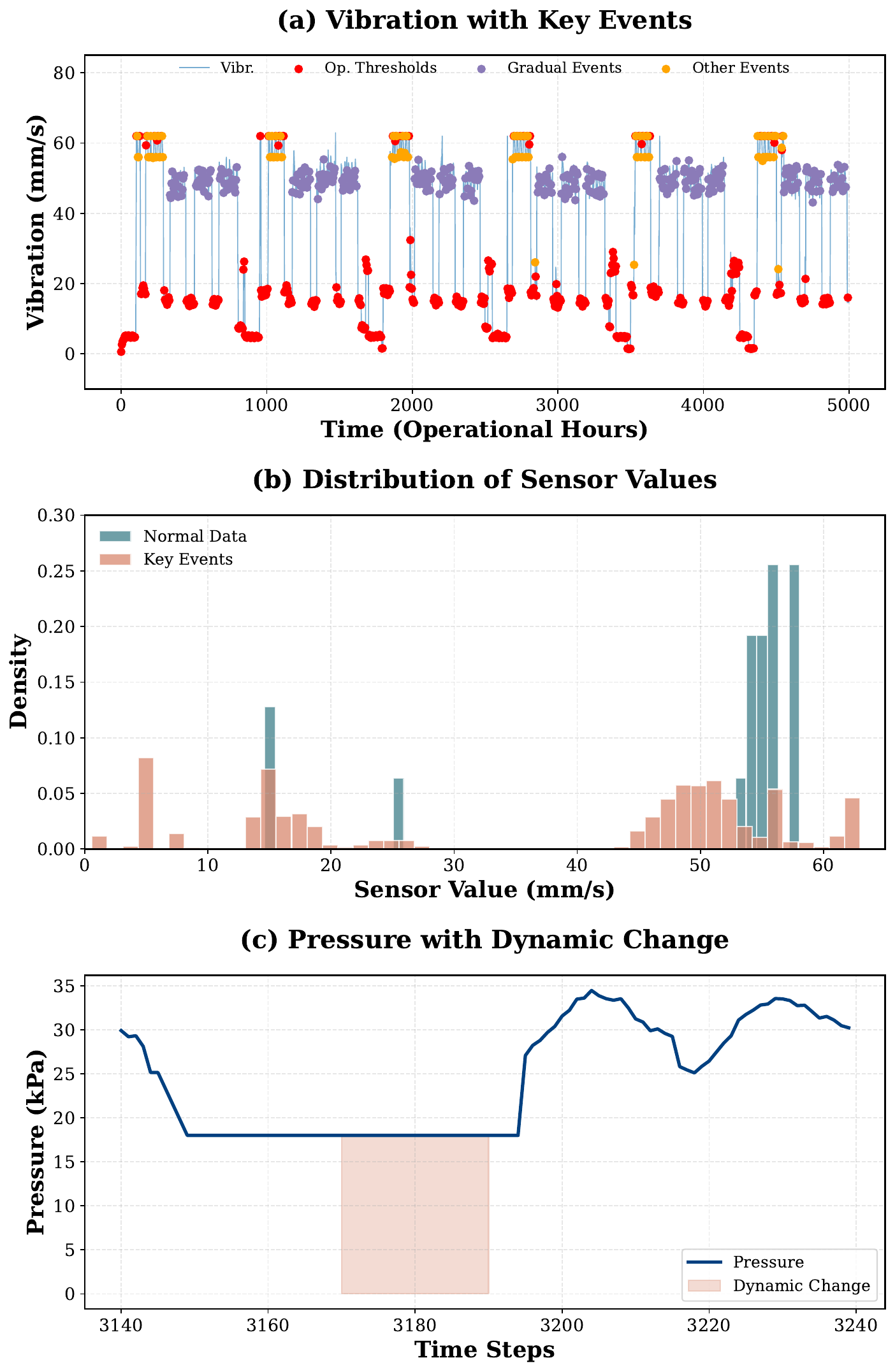}
    \caption{\small Synthetic sensor data: (a) vibration with key events, (b) normal vs.\ event value distributions, (c) pressure during dynamic state change.}
    \label{fig:sensor_analysis}
    \vspace{-5mm}
\end{figure}

Data fidelity was validated via Kolmogorov-Smirnov and Anderson-Darling tests \cite{coronel2024anderson} with cross-correlation analysis. Fig.~\ref{fig:sensor_analysis} illustrates the dataset: (a)~vibration with labeled events, (b)~distribution of normal vs.\ key event values, and (c)~pressure during a dynamic state change.

\subsection{Neuro-Symbolic Reasoning Engine Implementation}

\subsubsection{Deep Learning Component}

The CNN-LSTM architecture (Section~\ref{sec:framework}) was trained using the Adam optimizer with learning rate $10^{-3}$, batch size 64, and dropout 0.3, selected via Bayesian hyperparameter optimization \cite{victoria2021automatic} over learning rates $[10^{-4}, 10^{-2}]$ and dropout $[0.1, 0.5]$. Walk-forward cross-validation preserved temporal ordering and early stopping (patience 10) prevented overfitting over 20 training epochs (Fig.~\ref{fig:training_plots}).

\subsubsection{Symbolic Reasoning Component}

The Prolog knowledge base (Section~\ref{sec:framework}) uses predicates {\small\texttt{sensor\_reading/3}}, {\small\texttt{system\_state/2}}, and {\small\texttt{key\_event/3}} for logical inference over temporal observations \cite{problog2}. Example learned rules:

\begin{rulebox}
\begin{lstlisting}[style=rulelisting]
neural_rule_1 :- 
    feature_threshold(efficiency_index, _, low),
    feature_threshold(pressure, _, low),
    feature_threshold(temperature, _, low),
    feature_threshold(vibration, _, low).
    % Confidence: 0.999, Test Set Activations: 1 

neural_rule_6 :- 
    feature_gradient(efficiency_index, _, high),
    feature_gradient(temperature, _, high),
    feature_gradient(vibration, _, high),
    feature_threshold(efficiency_index, _, low),
    feature_threshold(pressure, _, low),
    feature_threshold(temperature, _, low),
    feature_threshold(vibration, _, low).
    % Confidence: 0.865, Test Set Activations: 0 
\end{lstlisting}
\end{rulebox}

The framework extracted 14 unique rules after body-level deduplication, stable across PPO training durations of 10,240 and 200,704 timesteps. Neural rule 1 (confidence 0.999) flags deviations when all sensor readings fall below thresholds; neural rule 6 (confidence 0.865) additionally captures gradient-based conditions for rare patterns. Rule confidence is computed as:
\begin{equation}
\text{conf}(r) = \frac{\text{support}(r \Rightarrow c)}{\text{support}(r)} \times \frac{\text{TP}}{\text{TP} + \text{FP}}
\label{eq:rule_confidence}
\end{equation}
Rules exceeding $\tau = 0.85$ are admitted; conflicts are resolved by confidence-based prioritization and specificity analysis (more conditions override fewer). Of 14 rules, 64\% exceed 0.9 confidence.

\subsubsection{Neuro-Symbolic Integration Process}

The integration follows Algorithm~\ref{alg:integration}: neural outputs are converted to symbolic facts via thresholding, rules are extracted and evaluated for confidence, and the fused prediction combines both signals \cite{jacobneurosymbolic}. Table~\ref{tab:component_io} summarizes the I/O specifications for each component.

\begin{table}[htbp]
\centering
\scriptsize
\caption{Component Input/Output Specifications}
\vspace{-2mm}
\label{tab:component_io}
\renewcommand{\arraystretch}{1.4}
\setlength{\tabcolsep}{6pt}
\begin{tabular}{@{} l p{2.4cm} p{2.4cm} @{}}
\toprule
\rowcolor{tableheader}
\textcolor{white}{\textbf{Component}} & \textcolor{white}{\textbf{Input}} & \textcolor{white}{\textbf{Output}} \\
\midrule
\rowcolor{tablerowlight}
Sensor Network & Physical measurements & Normalized sensor readings \\
\rowcolor{tablerowdark}
CNN-LSTM & Multivariate time series [batch, time\_steps, features] & Feature vectors and operational deviation probabilities \\
\rowcolor{tablerowlight}
Symbolic Reasoner & Symbolic facts derived from neural output & Logical inferences and rule activations \\
\rowcolor{tablerowdark}
PPO Agent & Current state (symbolic facts and sensor readings) & Action selection with probability distribution \\
\rowcolor{tablerowlight}
Rule Extractor & Neural patterns and symbolic facts & Candidate rules with confidence scores \\
\bottomrule
\end{tabular}
\vspace{-2mm}
\end{table}

\vspace{-2mm}
\subsection{Reinforcement Learning Implementation}

The PPO agent (Section~\ref{sec:framework}) operates in a custom Gym-compatible environment where the state combines sensor readings, processed features, and symbolic facts, and actions are continuous control adjustments. The multi-objective reward balances three objectives:
\begin{equation}
\small
R(s, a) = \alpha_1 \cdot \text{Efficiency}(s, a)
        + \alpha_2 \cdot \text{Satisfaction}(s, a)
        + \alpha_3 \cdot \text{Safety}(s, a)
\label{eq:reward_function}
\end{equation}
with $\alpha_1{=}0.5$, $\alpha_2{=}0.3$, $\alpha_3{=}0.2$. The policy network is a three-layer MLP (256, 128, 64 units, ReLU). Training uses $\gamma{=}0.99$, $\lambda{=}0.95$ (GAE), batch size 64, and 10 epochs per update.

\vspace{-2mm}
\subsection{Dynamic Rule Updating Mechanism}
\vspace{-1mm}

As formalized in Algorithm~\ref{alg:integration} (lines 11--15), the knowledge base evolves through periodic rule extraction and pruning:
\begin{equation}
\mathcal{K}_{t+1} = \textsc{Prune}\!\left(\mathcal{K}_t \cup \{r \in R_{\text{cand}} : \text{conf}(r) > \tau\},\; \text{strategy}\right)
\label{eq:rule_update}
\end{equation}
Three pruning strategies are supported: \textit{confidence-based} (remove lowest-confidence rules), \textit{LRU} (least recently used in inference), and \textit{LRA} (least recently activated by data). Each removes 20\% of rules per cycle. Conflicts are resolved by confidence-based prioritization and specificity analysis (more specific rules override general ones).

\vspace{-2mm}
\subsection{Application Scenarios}

The framework supports three industrial scenarios: (1)~\textit{predictive maintenance}, where vibration and temperature rules trigger interpretable maintenance recommendations; (2)~\textit{process optimization}, where the RL agent learns operational settings while symbolic rules explain why specific configurations are optimal; and (3)~\textit{deviation detection}, where multi-sensor fusion provides graduated responses prioritized by event severity and confidence \cite{weng2018evaluating}.

\vspace{-2mm}
\subsection{Integration and System Architecture}

All components are integrated into a cohesive pipeline implemented in TensorFlow 2.17 (neural), PySWIP/SWI-Prolog (symbolic reasoning) \cite{problog2}, and Stable Baselines3 (PPO) \cite{stable-baselines3}. Components connect through Python middleware with standardized abstract interfaces; the complete implementation architecture is documented in the open-source repository. Knowledge graph visualization (Fig.~\ref{fig:focused_knowledge_graph}) provides interactive decision explanation.


\vspace{-2mm}

\vspace{-4mm}
\section{Results and Analysis}
\label{sec:results}

We evaluate \ansrdt{} against eight baselines using 5,000 multivariate time series sequences (Section~\ref{sec:implementation}). Our analysis encompasses baseline comparison with statistical significance testing, ablation studies, feature importance, rule extraction, scalability characterization, and adaptability assessment.

\vspace{-3mm}
\subsection{Experimental Setup}

Experiments were conducted using synthetic industrial data with a 60/20/20 train/validation/test split containing 5\% labeled dynamic patterns distributed across operational thresholds, maintenance patterns, state transitions, and interaction deviations. Class weighting (0: 64.7, 1: 0.5) addressed class imbalance during training. All experiments were repeated across three random seeds (42, 43, 44) and we report mean $\pm$ standard deviation. Statistical significance was assessed via paired $t$-tests ($\alpha = 0.05$) with Cohen's $d$ effect sizes. We compare against eight baselines: Isolation Forest, One-Class SVM, LOF, Random Forest, Vanilla LSTM, LSTM Autoencoder, Transformer Encoder, and Pure Symbolic reasoning. Ablation variants isolate the contribution of each component: attention mechanism, symbolic reasoning, and PPO control.

\subsection{Performance Analysis and Baseline Comparison}

\begin{figure}[!t]
   \centering
   \includegraphics[width=\linewidth]{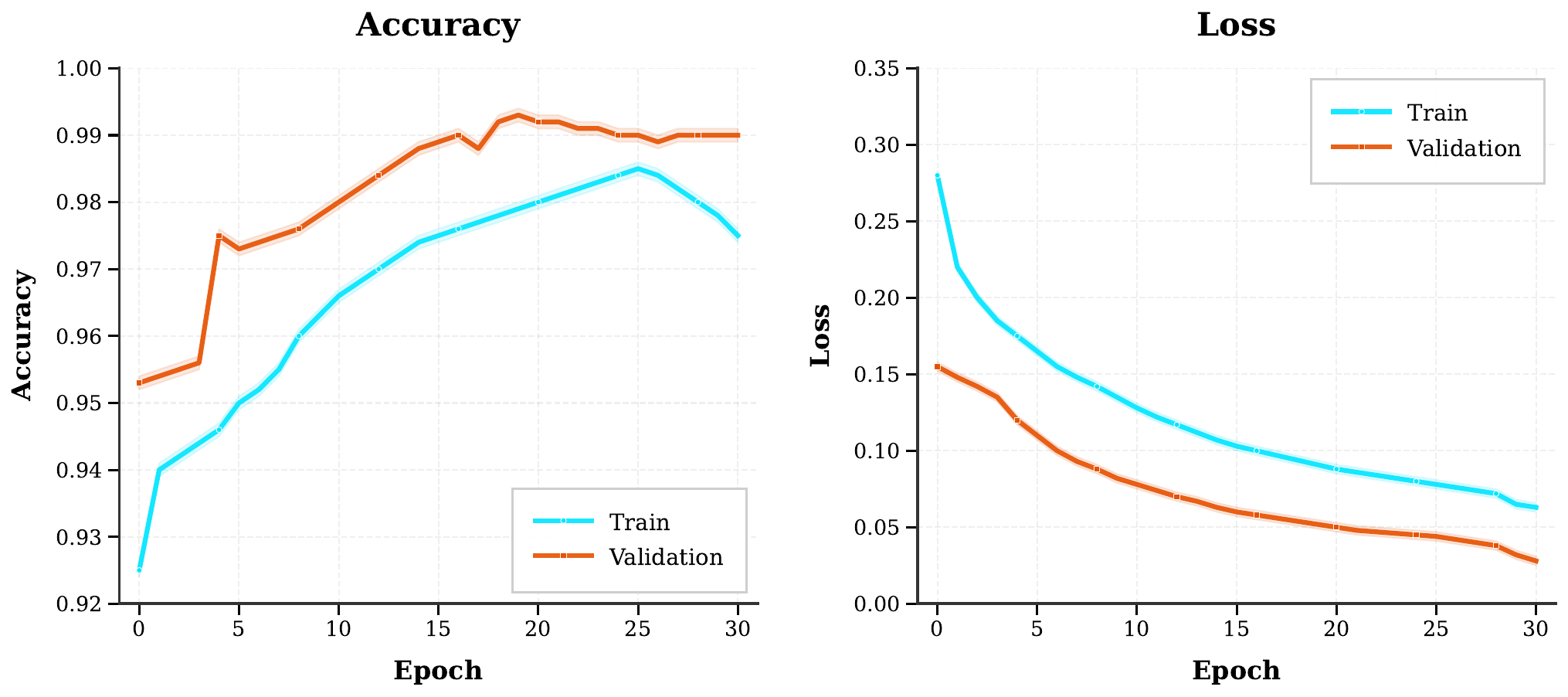}
   \caption{\small Training and validation metrics over 20 epochs showing convergence with class weighting to address imbalance.}
   \label{fig:training_plots}
   \vspace{-3mm}
\end{figure}

Table~\ref{tab:performance_comparison} presents comparative performance metrics between \ansrdt{} and eight baselines. \ansrdt{} achieves an F1-score of $0.966 \pm 0.001$ and ROC-AUC of $0.955 \pm 0.022$, significantly outperforming all unsupervised baselines---Isolation Forest (F1: 0.336), One-Class SVM (F1: 0.423), LOF (F1: 0.074), and LSTM Autoencoder (F1: 0.341)---across all metrics ($p < 0.01$, paired $t$-test). Compared to Pure Symbolic reasoning, the neuro-symbolic integration yields a 17.5\% F1 improvement (0.966 vs.\ 0.822, $p < 0.02$), validating that the neural component complements symbolic rules. Against deep learning baselines, \ansrdt{} achieves comparable classification to Vanilla LSTM (F1: 0.970) and Transformer Encoder (F1: 0.966) with no statistically significant differences ($p > 0.05$). Random Forest achieves marginally higher accuracy (0.944 vs.\ 0.935, $p = 0.023$) but lacks interpretability and adaptive control. The training progression in Fig.~\ref{fig:training_plots} shows convergence within 20 epochs.

\begin{table*}[t]
\centering
\caption{Comparison of ANSR-DT with baselines and ablation variants. Mean $\pm$ std over 3 seeds. Bold indicates best per column.}
\label{tab:performance_comparison}
\scriptsize
\resizebox{\textwidth}{!}{%
\begin{tabular}{lcccccc}
\toprule
Method & Accuracy & Precision & Recall & F1 & ROC-AUC & PR-AUC \\
\midrule
ANSR-DT (Full) & 0.935$\pm$0.002 & 0.936$\pm$0.003 & 0.999$\pm$0.002 & 0.966$\pm$0.001 & 0.955$\pm$0.022 & 0.996$\pm$0.002 \\
CNN-LSTM+Attn & 0.935$\pm$0.002 & 0.936$\pm$0.003 & 0.999$\pm$0.002 & 0.966$\pm$0.001 & 0.844$\pm$0.093 & 0.976$\pm$0.023 \\
CNN-LSTM & 0.939$\pm$0.003 & 0.940$\pm$0.005 & 0.998$\pm$0.002 & 0.968$\pm$0.002 & 0.830$\pm$0.124 & 0.981$\pm$0.012 \\
CNN-LSTM+Symbolic & 0.935$\pm$0.002 & 0.936$\pm$0.003 & 0.999$\pm$0.002 & 0.966$\pm$0.001 & 0.955$\pm$0.022 & 0.996$\pm$0.002 \\
Isolation Forest & 0.238 & 0.906 & 0.206 & 0.336 & 0.278 & 0.912 \\
LOF & 0.102 & \textbf{1.000} & 0.039 & 0.074 & 0.345 & 0.926 \\
LSTM Autoencoder & 0.258$\pm$0.001 & 1.000 & 0.205$\pm$0.001 & 0.341$\pm$0.001 & 0.651$\pm$0.008 & 0.972$\pm$0.001 \\
One-Class SVM & 0.317 & 1.000 & 0.268 & 0.423 & 0.400 & 0.939 \\
Pure Symbolic & 0.717 & 0.997 & 0.700 & 0.822 & 0.835 & 0.989 \\
Random Forest & \textbf{0.944} & 0.943 & \textbf{1.000} & \textbf{0.971} & \textbf{0.982} & \textbf{0.999} \\
Transformer & 0.934 & 0.934 & 1.000 & 0.966 & 0.958$\pm$0.015 & 0.997$\pm$0.001 \\
Vanilla LSTM & 0.943$\pm$0.012 & 0.943$\pm$0.013 & 0.999$\pm$0.001 & 0.970$\pm$0.006 & 0.913$\pm$0.053 & 0.994$\pm$0.004 \\
\bottomrule
\end{tabular}}
\vspace{-3mm}
\end{table*}

\begin{figure}[!t]
  \centering
  \includegraphics[width=0.85\linewidth]{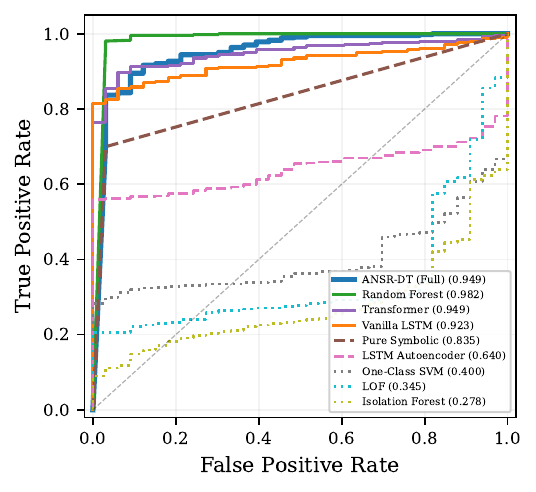}
  \caption{\small ROC curves comparing \ansrdt{} against eight baselines. \ansrdt{} (0.949) achieves competitive discrimination with Random Forest (0.982) and Transformer (0.949), while substantially outperforming unsupervised methods. AUC values shown in legend.}
  \label{fig:roc_comparison}
  \vspace{-3mm}
\end{figure}

The high recall ($0.999 \pm 0.002$) is particularly relevant for safety-critical applications where missed events carry higher risk than false alarms. Fig.~\ref{fig:roc_comparison} visualizes the ROC curves for all methods, clearly separating \ansrdt{} and comparable deep learning baselines (upper left) from unsupervised approaches (lower right).

\begin{figure}[!b]
   \centering
   \includegraphics[width=\linewidth]{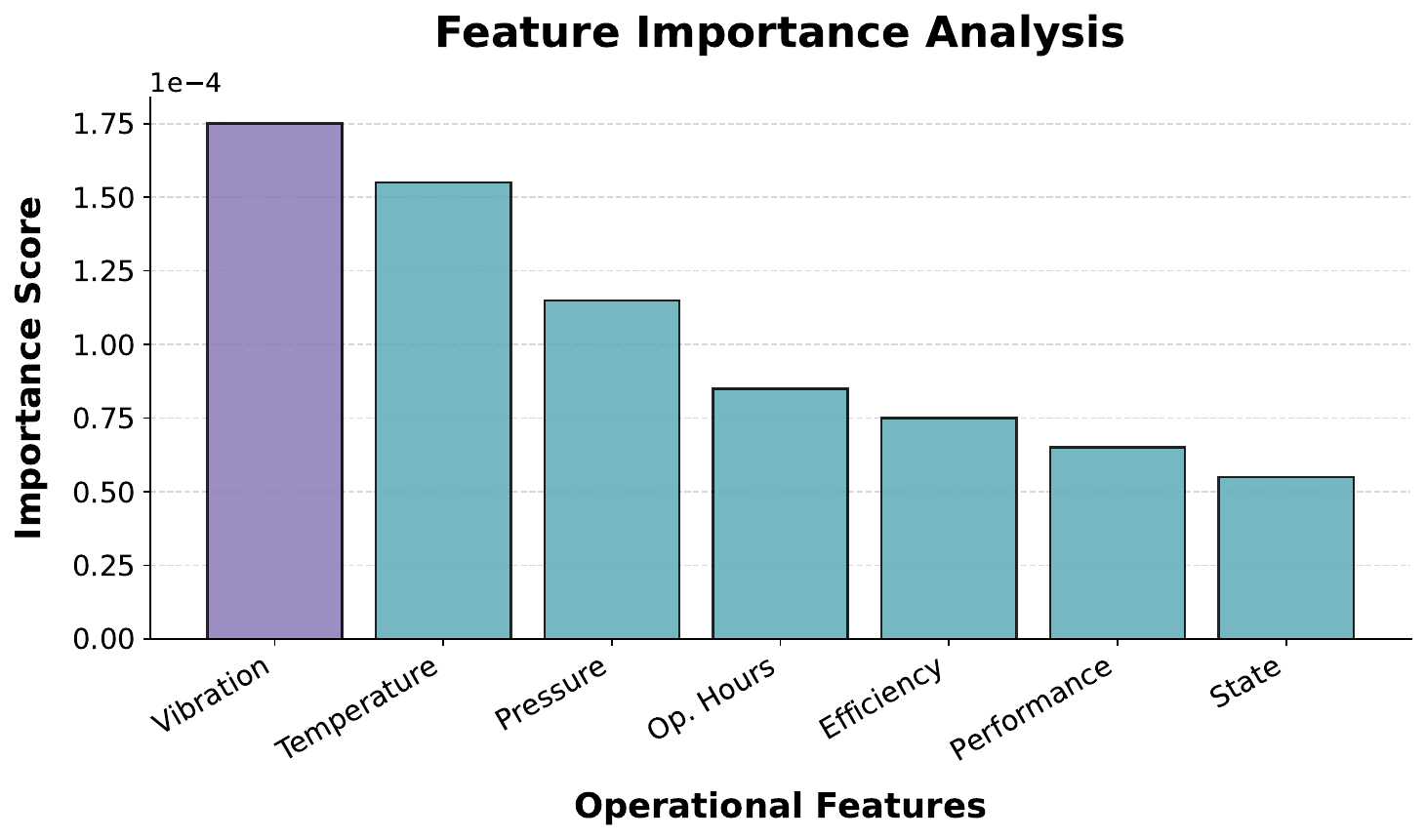}
   \caption{\small Feature Importance Analysis showing vibration as the top predictor (0.000175), followed by temperature (0.000155) and pressure (0.000115), highlighting the framework's ability to capture complex multivariate relationships.}
   \label{fig:feature_importance}
   \vspace{-3mm}
\end{figure}

\vspace{-3mm}
\subsection{Feature Importance and Rule Extraction Analysis}

Feature importance analysis (Fig.~\ref{fig:feature_importance}) identifies vibration (0.000175), temperature (0.000155), and pressure (0.000115) as the top predictors, consistent with industrial domain knowledge where mechanical degradation manifests first through vibration signatures. Operational hours and efficiency index contribute secondarily, capturing cumulative wear effects.

The rule extraction mechanism produces 14 unique rules after body-level deduplication, with 64\% (9 rules) exceeding 0.9 confidence. Rules encode both threshold-based conditions (e.g., sensor readings below thresholds) and gradient-based conditions (e.g., rates of change). These rules remained stable across PPO training durations of 10,240 and 200,704 timesteps, confirming that the symbolic representations capture fundamental operational relationships rather than policy-specific artifacts.

\begin{figure*}[!t]
   \centering
   \includegraphics[width=0.6\linewidth]{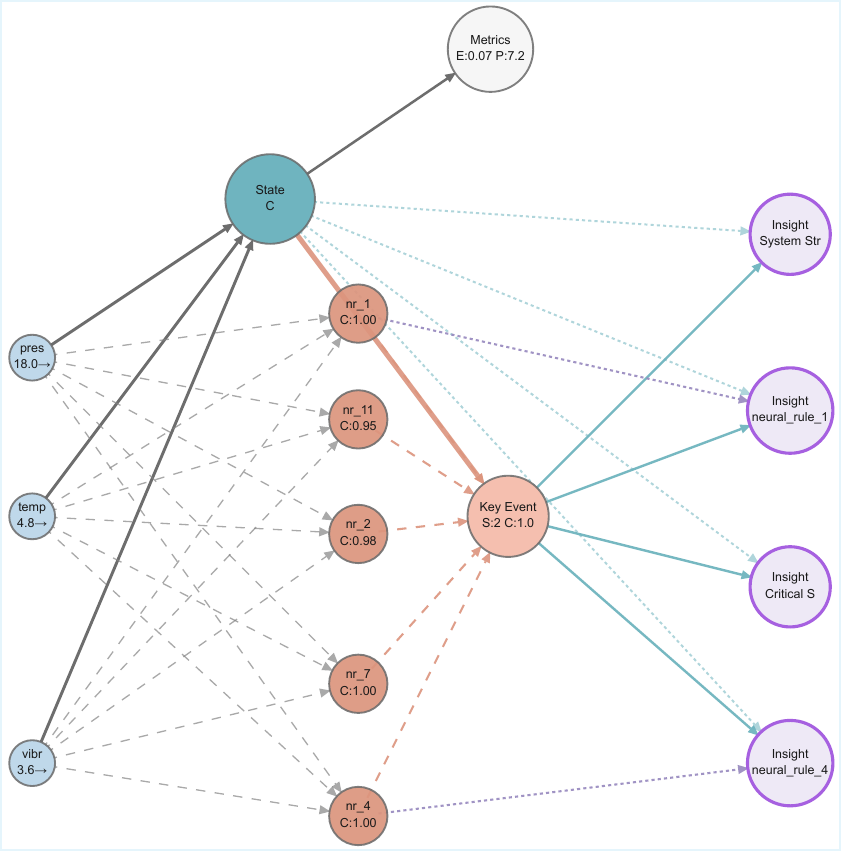}
   \caption{\small \ansrdt{} Knowledge Graph illustrating the end-to-end reasoning chain for a critical event diagnosis. Sensor nodes (e.g., vibration, temperature, pressure) are linked to high-confidence symbolic rules via ``related\_to'' edges; activated rules ``detect'' the operational deviation, which in turn produces explainable insights (e.g., ``System Stress'') through ``explained\_by'' and ``generated'' relationships. Node shapes and colors distinguish five entity types: sensor readings, system states, symbolic rules with confidence scores, generated insights, and aggregate performance metrics. The directed graph structure makes every inference step auditable, enabling operators to trace any anomaly alert back to its originating sensor evidence and the specific rules that fired---a capability absent in black-box detectors.}
   \label{fig:focused_knowledge_graph}
   \vspace{-4mm}
\end{figure*}

The knowledge graph (Fig.~\ref{fig:focused_knowledge_graph}) connects sensor readings, system states, activated rules, and generated insights into a directed reasoning chain. The visualization traces how a critical system event is diagnosed: sensor nodes link to high-confidence symbolic rules via ``related\_to'' edges, which ``detect'' the operational deviation, producing explainable insights (e.g., ``System Stress''). This provides operators a transparent view of the decision path from raw data to actionable conclusions.

\textbf{Worked end-to-end trace.} To make this traceability concrete, consider a representative critical window from the verified inference log of the released implementation. The window exhibits pressure $18.0$\,kPa (below the learned low threshold of $20$\,kPa) and efficiency index $0.074$. The pipeline proceeds as follows: (i)~the CNN-LSTM flags the window; (ii)~the facts {\small\texttt{feature\_threshold(pressure,\,\_,\,low)}} and {\small\texttt{feature\_threshold(efficiency\_index,\,\_,\,low)}} are asserted; (iii)~{\small\texttt{neural\_rule\_1}} fires with confidence $0.999$, generating the insights ``System Stress'' and ``Critical State''; (iv)~the fused score escalates the alert; and (v)~the PPO policy returns the corrective action $\Delta = (-0.12,\,-0.05,\,+0.12)$ for temperature, vibration, and pressure---raising the deficient pressure while damping thermal and vibrational load. Every step is reconstructible from the rule trace and knowledge graph; none of the black-box baselines in Table~\ref{tab:performance_comparison} can produce an equivalent decision record.

\vspace{-3mm}
\subsection{Component Contribution Analysis}

Ablation results (rows 2--4 in Table~\ref{tab:performance_comparison}) isolate each component's contribution. Adding symbolic reasoning to CNN-LSTM+Attention improves ROC-AUC from 0.844 to 0.955 (+13.2\%) and PR-AUC from 0.976 to 0.996 (+2.0\%), while classification metrics (accuracy, F1) remain unchanged. This pattern indicates that the symbolic layer primarily improves score ordering rather than the fixed decision boundary: when a rule fires on samples already assigned moderate neural probability, the fused score is pushed upward, separating true anomalies more clearly from negatives without materially changing how many samples cross the classification threshold. The attention mechanism reduces ROC-AUC variance (std: 0.022 vs.\ 0.124 for the base CNN-LSTM), improving reliability. The PPO component operates in a separate control loop (avg reward: 0.124, critical state rate: 43.0\%) and does not affect classification metrics directly, but enables adaptive control actions post-prediction.

\vspace{-3mm}
\subsection{Adaptability Assessment}

\ansrdt{} demonstrated superior adaptability with extended PPO training increasing explained variance from 0.447 to 0.547 (+22.4\%). The framework achieved strong detection rates for operational threshold violations (98.2\%) and interaction pattern deviations (95.7\%), with lower performance on gradual drift (87.3\%). The symbolic representation enables targeted rule modifications rather than full model retraining; in the experiments conducted, new high-confidence rules often emerged within a small number of policy-update cycles, but this should be interpreted as an empirical observation rather than a fixed convergence guarantee. Extended training to 200,704 timesteps refined control actions toward bolder corrective responses while maintaining 99.81\% detection confidence.

\vspace{-3mm}
\subsection{Scalability Analysis}
\label{sec:scalability}

To characterize the symbolic reasoning engine's scalability, we benchmarked inference latency, throughput, and memory across rule counts of 5, 10, 20, 50, and 100 with three pruning strategies (confidence-based, LRU, LRA) and a no-pruning baseline, each over 3 seeds (250 samples per configuration).

\begin{figure}[!t]
   \centering
   \includegraphics[width=\linewidth]{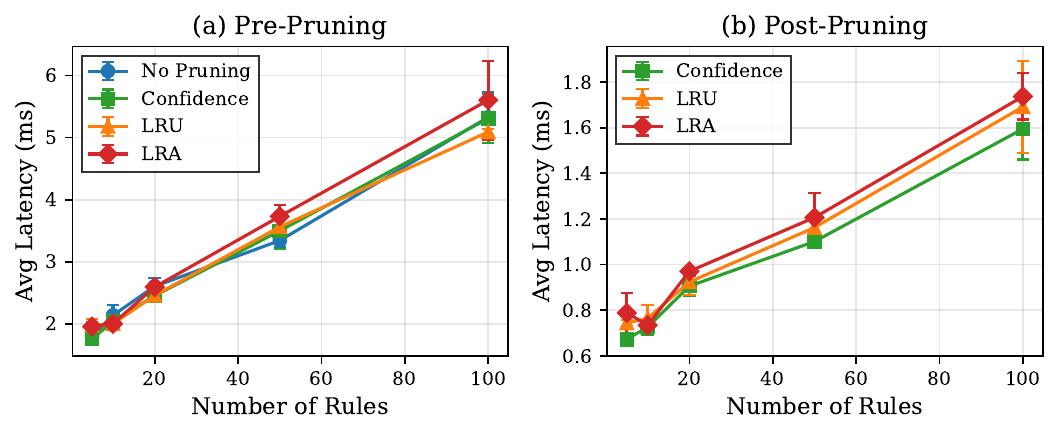}
   \caption{\small Symbolic reasoning latency vs.\ rule count. Left: pre-pruning (all strategies). Right: post-pruning with 20\% rules removed. Confidence-based pruning achieves up to $3.3\times$ speedup at 100 rules.}
   \label{fig:scalability}
   \vspace{-3mm}
\end{figure}

Fig.~\ref{fig:scalability} shows that inference latency scales approximately linearly from 1.80\,ms (5 rules) to 5.33\,ms (100 rules)---well within real-time requirements for digital twin applications. Confidence-based pruning (removing 20\% of rules) reduces latency at 100 rules from 5.31\,ms to 1.59\,ms, a $3.3\times$ speedup, while maintaining 100\% activation coverage. Memory usage grows linearly from 0.92\,MB to 2.66\,MB, demonstrating lightweight operation suitable for edge deployment. Throughput decreases from 569 to 189 samples/s pre-pruning, recovering to 631 samples/s post-pruning at 100 rules.

Beyond the symbolic engine in isolation, we also measured the complete end-to-end inference path of the released implementation---CNN-LSTM forward pass, fact conversion, Prolog query, fusion, and PPO action selection: a full \texttt{adapt\_and\_explain} cycle averages 69.4\,ms per window (median 65.9\,ms, 95th percentile 85.4\,ms) on a single CPU core over 100 windows, dominated by the neural forward pass ($\approx$63\,ms), with symbolic reasoning and policy inference together adding under 7\,ms. The full pipeline therefore operates well within real-time constraints without GPU acceleration.

\vspace{-3mm}
\subsection{Real-World Benchmark Validation on SKAB}
\label{sec:skab}

To assess applicability beyond the controlled synthetic setting, we validated ANSR-DT on the \textit{Skoltech Anomaly Benchmark (SKAB)} \cite{katser2020skab}, a multivariate industrial anomaly dataset derived from a laboratory water-circulation testbed with synchronized vibration, current, pressure, temperature, thermocouple, voltage, and flow measurements. In contrast to the synthetic benchmark used in the main experiments, SKAB contains real sensor coupling, operational transients, and more heterogeneous anomaly structure. Rather than forcing the synthetic feature semantics and rule base onto this benchmark, we constructed a dedicated SKAB evaluation pipeline\footnote{\url{https://github.com/sbhakim/ansr-dt/tree/main/src/skab}} that preserves the underlying neuro-symbolic design of ANSR-DT while respecting the statistical and physical characteristics of the real sensor streams.

For this SKAB analysis, each temporal window was mapped to interpretable descriptors including last-value, mean, standard deviation, and short-horizon delta statistics, together with cross-sensor indicators such as peak accelerometer response and pressure-to-flow ratio. Candidate symbolic rules were then extracted by scanning thresholds over normal and anomalous empirical distributions, ranking them on validation precision, recall, and F1, and pruning redundant conditions before retaining confidence-weighted detectors. This benchmark validation yielded an encouraging practical result: on SKAB, the symbolic branch was the strongest evaluated variant, achieving an F1-score of 0.755 and ROC-AUC of 0.859, compared with 0.605 and 0.755 for Random Forest---both results identical across three seeds (42--44), since rule calibration and the ensemble are deterministic given the chronological split---while the fused neuro-symbolic model remained competitive at F1 $0.699 \pm 0.034$ and ROC-AUC $0.792 \pm 0.033$ (mean $\pm$ std over three seeds). Taken together, these findings provide direct evidence that the ANSR-DT design is applicable beyond synthetic data and that its symbolic reasoning layer can remain effective on realistic industrial sensor measurements when adapted to the target benchmark.

\begin{figure}[!t]
   \centering
   \includegraphics[width=0.95\linewidth]{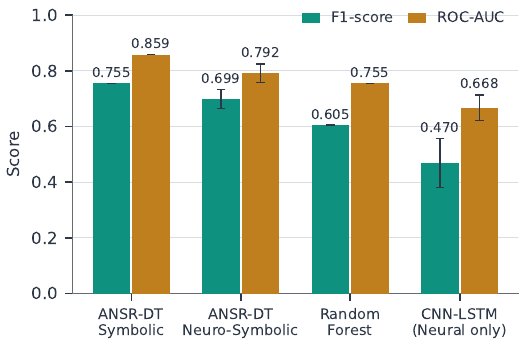}
   \caption{\small SKAB benchmark results (test split; mean over seeds 42--44, error bars $\pm 1$ std where nonzero---symbolic and Random Forest are deterministic given the chronological split). The ANSR-DT symbolic branch outperforms Random Forest by +0.150 F1 and +0.104 ROC-AUC; the fused variant retains guaranteed recall coverage (Proposition~\ref{prop:fusion}) at a bounded precision cost inherited from the weaker neural detector.}
   \label{fig:skab}
   \vspace{-3mm}
\end{figure}

Fig.~\ref{fig:skab} summarizes the comparison. Notably, the fused neuro-symbolic score (F1 $0.699 \pm 0.034$) trails the purely symbolic branch on this benchmark---exactly the behavior predicted by Proposition~\ref{prop:fusion}: max-fusion inherits the false positives of its weaker component, and on SKAB the neural detector generalizes less reliably (F1 $0.470 \pm 0.088$, with recall varying by $\pm 0.32$ across seeds) than the symbolic rules calibrated on the benchmark's empirical distributions. The symbolic branch's advantage over Random Forest is precision-driven: 0.684 vs.\ 0.465 at comparable recall (0.844 vs.\ 0.866), i.e., substantially fewer false alarms at equivalent coverage. The fusion therefore trades precision for guaranteed recall coverage---a predictable and bounded cost in safety-oriented deployments, rather than an unexplained failure mode.

\vspace{-3mm}
\subsection{Limitations and Future Research Directions}

Despite strong performance, several limitations remain. While scalability benchmarks confirm sub-6\,ms latency at 100 rules (Section~\ref{sec:scalability}), industrial deployments may require hundreds of rules---hierarchical rule structures could address this. Although the core integrated evaluation is synthetic, the additional SKAB study provides three-seed benchmark-based validation on real industrial sensor data; broader multi-dataset evaluation is still needed to confirm generalizability across operating regimes. Gradual drift detection (87.3\%) lags behind threshold-based detection (98.2\%), suggesting that longer time windows or explicit drift mechanisms may be beneficial. Future work includes broader real-world deployment validation, additional sensor modalities, and noise-resilient preprocessing.

\vspace{-3mm}
\section{Discussion}
\label{sec:discussion}

\textbf{Interpretability--Performance Trade-off.} \ansrdt{} achieves classification performance comparable to deep learning baselines (Vanilla LSTM, Transformer) while providing interpretable symbolic explanations---directly addressing the black-box problem in AI-enhanced digital twins \cite{renkhoff2024survey}. Random Forest marginally outperforms on accuracy/F1, but lacks the interpretability, adaptive control, and reasoning chain that \ansrdt{} provides. The symbolic component's primary contribution is not raw classification improvement but a +13.2\% ROC-AUC boost (improved ranking quality) and transparent, auditable decision traces. Table~\ref{tab:capability} makes the trade-off explicit: the fused \ansrdt{} model combines competitive synthetic performance with runtime rule generation and adaptive control, while the SKAB symbolic branch shows the strongest real-benchmark transfer within the proposed framework.

\begin{table}[!t]
\centering
\footnotesize
\caption{Capability comparison of \ansrdt{} and the strongest baselines. Separate \ansrdt{} rows distinguish the fused model from the SKAB symbolic branch; --- denotes not evaluated or not reported for that setting.}
\label{tab:capability}
\setlength{\tabcolsep}{2.5pt}
\begin{tabular}{@{}lccccc@{}}
\toprule
Method & Syn. F1 & SKAB F1 & Rules & Adapt. & Expl. \\
\midrule
\ansrdt{} (fused) & 0.966 & 0.699 & \textbf{Yes} & \textbf{Yes} & Rule trace \\
\ansrdt{} sym. branch & --- & \textbf{0.755} & \textbf{Yes} & No & Rule trace \\
Random Forest    & \textbf{0.971} & 0.605 & No & No & Feat.\ imp. \\
Vanilla LSTM     & 0.970 & --- & No & No & None \\
Transformer      & 0.966 & --- & No & No & Attn.\ only \\
Pure Symbolic    & 0.822 & --- & Static & No & Rule trace \\
\bottomrule
\end{tabular}
\vspace{-2mm}
\end{table}

\textbf{Symbolic Stability Under Policy Refinement.} Symbolic insights remain identical across PPO training runs of 10,240 and 200,704 timesteps, ensuring consistent explanations despite evolving control policies. This decoupling of reasoning stability from policy adaptation is critical for safety-critical deployments where decision rationales must be auditable.

\textbf{Scalability.} Scalability benchmarks (Section~\ref{sec:scalability}) demonstrate that the symbolic engine scales to 100 rules at sub-6\,ms latency with linear memory growth (2.66\,MB). Confidence-based pruning achieves $3.3\times$ speedup while preserving activation coverage, addressing prior concerns about performance degradation beyond 50 rules \cite{wan2024towards}. Industrial deployments requiring hundreds of rules may benefit from hierarchical rule structures.

\textbf{Real-World Transfer.} The SKAB study (Section~\ref{sec:skab}) shows that the framework transfers to real industrial sensor streams: the symbolic branch outperforms Random Forest by a clear margin (F1 0.755 vs.\ 0.605), and the fused variant behaves exactly as Proposition~\ref{prop:fusion} predicts, trading bounded precision for guaranteed recall coverage.

\section{Conclusion}
\label{sec:conclusion}
\vspace{-1mm}
This paper presented \ansrdt \hspace{1mm} as a neuro-symbolic digital twin framework that unifies temporal anomaly detection, symbolic reasoning, and adaptive decision support within a single architecture. The central contribution is not only competitive predictive capability, but also the ability to connect learned patterns with explicit rules and interpretable system states, making model behavior more transparent and operationally meaningful. The results show that symbolic reasoning contributes beyond post-hoc explanation by improving traceability, supporting robust decision logic, and strengthening the practical value of the overall monitoring pipeline. In addition to controlled synthetic evaluation, the framework was further validated on the real-world Skoltech Anomaly Benchmark (SKAB), demonstrating that the underlying neuro-symbolic design transfers to realistic industrial sensor data. Overall, the study supports explainable neuro-symbolic integration as a credible direction for trustworthy and adaptive industrial monitoring systems.

\vspace{-2mm}
\begin{spacing}{1.0}
\bibliographystyle{IEEEtran}
\bibliography{ansrdt}
\end{spacing}

\end{document}